\pdfoutput=1

\documentclass[11pt]{article}

\usepackage[]{acl}

\usepackage{times}
\usepackage{latexsym}

\usepackage{adjustbox}
\usepackage{amsmath}
\usepackage{amssymb}
\usepackage{booktabs}
\usepackage{geometry}
\usepackage{listings}
\usepackage{multirow}
\usepackage{multicol}

\usepackage[T1]{fontenc}

\usepackage[utf8]{inputenc}

\usepackage{microtype}

\title{Small Models Are (Still) Effective Cross-Domain Argument Extractors}

\author{William Gantt \and Aaron Steven White \\
University of Rochester \\
\texttt{\{wgantt@cs.|aaron.white@\}rochester.edu}}

\begin{document}
\maketitle
\begin{abstract}
Effective ontology transfer has been a major goal of recent work on event argument extraction (EAE). Two methods in particular---question answering (QA) and template infilling (TI)---have emerged as promising approaches to this problem. However, detailed explorations of these techniques' ability to actually enable this transfer are lacking. In this work, we provide such a study, exploring zero-shot transfer using both techniques on six major EAE datasets at both the sentence and document levels. Further, we challenge the growing reliance on LLMs for zero-shot extraction, showing that vastly smaller models trained on an appropriate source ontology can yield zero-shot performance superior to that of GPT-3.5 or GPT-4.\footnote{\tiny{\url{https://github.com/wgantt/eae-transfer}}}
\end{abstract}

\section{Introduction}
\label{sec:intro}
\begin{figure}
    \centering
    \includegraphics[width=\columnwidth]{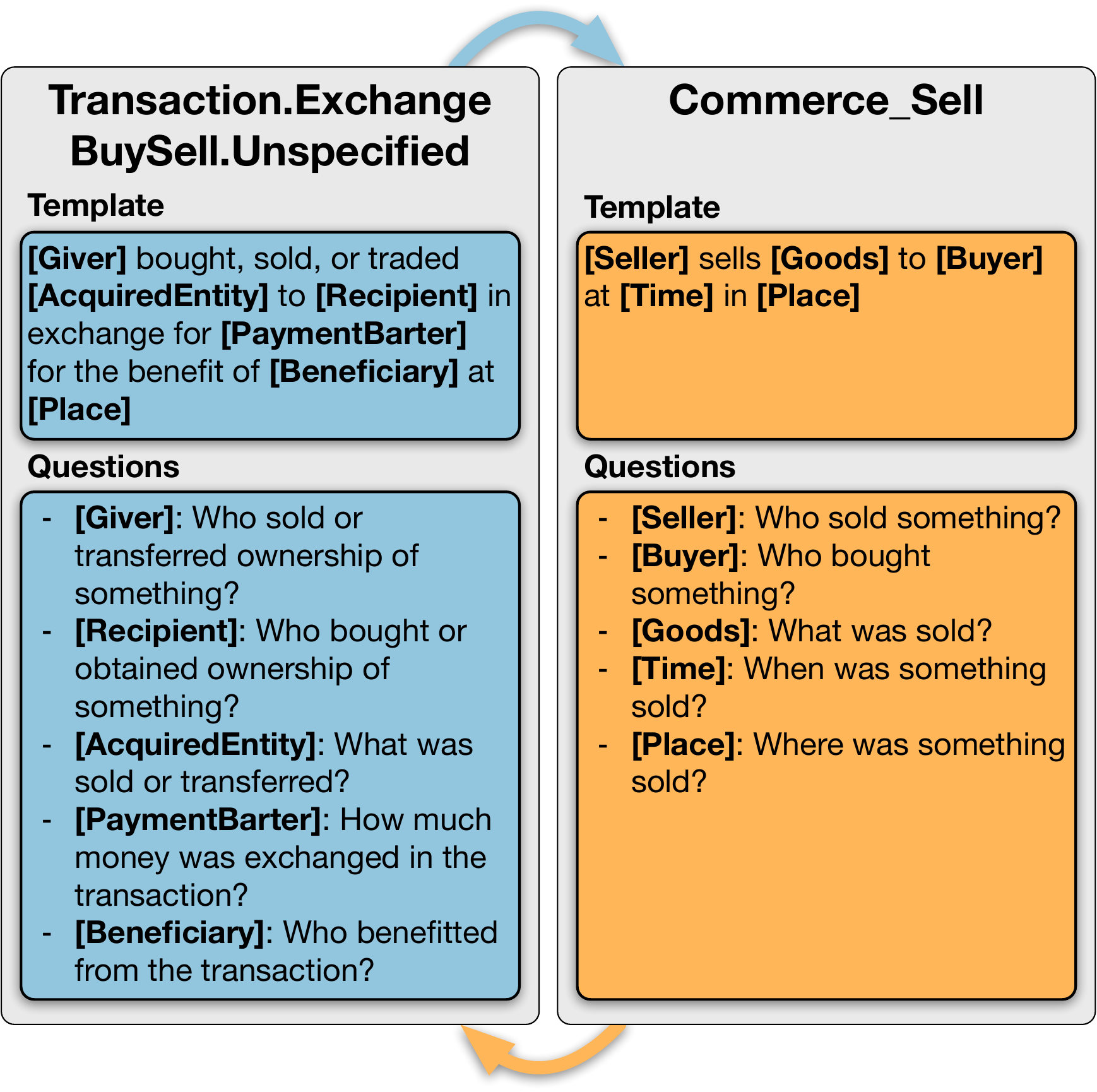}
    \caption{Example event types from WikiEvents (left) and from FAMuS (right), along with the templates and questions used for them in this work.\vspace{-5.5mm}}
    \label{fig:fig1}
\end{figure}

Traditional approaches to event argument extraction (EAE) involve training a model to identify and classify arguments against a fixed ontology, rendering zero-shot transfer to new ontologies impossible. However, recent works have proposed reformulations of the EAE task that in principle enable such transfer. Among these approaches, question answering \citep[QA;][\emph{i.a.}]{du-cardie-2020-event, liu-etal-2020-event, li-etal-2020-event} and template infilling \citep[TI;][\emph{i.a.}]{chen-etal-2020-reading, li-etal-2021-document, hsu-etal-2022-degree} have emerged as especially promising. In the former approach, role labels are recast as participant-focused questions, and in the latter, the full role set for a given event type is expressed in a templatic prompt to be filled with extracted arguments (\autoref{fig:fig1}). Handling new roles thus becomes a matter of writing new questions or templates.

Investigations of the effectiveness of zero-shot transfer using these methods have generally been limited either to only one or two ontology pairs \citep{zhang-etal-2022-transfer}, or to splits of event types within the same ontology \citep{li-etal-2021-document, du-cardie-2020-event}. This work broadens these investigations with the following contributions:
\begin{itemize}
    \item We study transfer with both TI and QA at a larger scale, benchmarking and analyzing transfer performance between six sentence- and document-level EAE datasets.
    \item We provide expert-written questions and templates for all roles in all six ontologies, which may be leveraged for transfer learning.
    \item We show that small QA and TI models based on Flan-T5 \citep{chung-etal-2022-scaling} can yield zero-shot performance superior to that of GPT-3.5 and GPT-4, given a suitable source ontology.
\end{itemize}

\section{Approaches}
\label{sec:approaches}
\paragraph{Question Answering (QA)} Many works have sought to recast EAE as a question answering (QA) task, in which questions correspond to specific roles (\autoref{fig:fig1}). Such questions range widely in form---from almost brute verbalizations of role names \citep[e.g.\ \emph{What is the Org?};][]{du-cardie-2020-event}, to semantically bleached queries with rule-generated syntax \citep[\emph{What was someone expected to do?};][]{he-etal-2015-question}, to detailed, expert-written questions \citep[\emph{What did this person/network do that was corrupt?};][]{holzenberger-etal-2022-asking}.

We imagine a typical use case for a QA-based EAE system as one in which a human user is able to write queries about roles of interest in fairly plain language. Accordingly, we (the authors) write simple questions for all roles in the six ontologies we consider, avoiding use of the actual role names in the question text. Our questions draw stylistic inspiration from QA-SRL \citep{he-etal-2015-question}, with the key difference that they are \emph{type-level}, and \emph{not} relativized to a predicate in context.\footnote{Having to write a new question for each predicate does not align with our envisioned use case.}

\paragraph{Template Infilling (TI)} Other works have treated EAE as an infilling task, in which slots in fixed templates for each event type are populated with extracted arguments, either via masked language modeling \citep{chen-etal-2020-reading} or via autoregressive generation of the filled template \citep{ma-etal-2022-prompt, li-etal-2021-document, huang-etal-2021-document, du-etal-2021-grit}. There are two notable advantages of this approach over QA: (1) for architectures with an encoder, only a single encoder forward pass is needed to extract all arguments, and (2) roles are considered jointly.

A template typically consists of a short, semi-natural description of its event type, with slots occupying the syntactic position that arguments for the corresponding role are expected to fill. We manually construct templates for all ontologies considered in this work. We follow \citet{ma-etal-2022-prompt} in representing slots by their associated role name (leveraging label semantics), and \citet{li-etal-2021-document} in coordinating multiple arguments for the same role with ``and.'' \autoref{fig:fig1} shows example templates.

\section{Experiments}
\label{sec:experiments}
\begin{table*}
    \centering
    \begin{tabular}{l|c|c|cccccc}
    \toprule
        Model & Method & Source Dataset & ACE & ERE-L & ERE-R & FAMuS & RAMS & WikiEvents \\
    \midrule
        \multirow{6}{*}{Flan-T5} & \multirow{6}{*}{TI} & ACE & \textbf{65.95} & 46.31 & 37.47 & 16.37 & 26.50 & 13.32 \\
        & & ERE-L & 32.88 & \textbf{66.71} & \underline{51.57} & 14.35 & 23.94 & 23.05 \\
        & & ERE-R & \underline{48.14} & 62.07 & \textbf{66.78} & 23.03 & 30.18 & 28.21 \\
        & & FAMuS & 32.34 & 28.89 & 24.87 & 43.81 & 24.50 & \phantom{0}8.75 \\
        & & RAMS & 32.20 & 34.40 & 26.97 & 17.71 & 48.47 & 25.37 \\
        & & WikiEvents & 25.66 & 31.72 & 32.25 & \phantom{0}8.04 & 29.29 & \textbf{64.20} \\
        \midrule
        \multirow{6}{*}{Flan-T5} & \multirow{6}{*}{QA} & ACE & 60.35 & 48.52 & 48.50 & 27.34 & 30.25 & 18.87 \\
        & & ERE-L & 31.47 & 63.96 & 44.59 & 22.25 & 29.96 & 24.81 \\
        & & ERE-R & 43.89 & \underline{62.34} & 66.00 & 28.47 & \underline{35.28} & 32.43 \\
        & & FAMuS & 34.83 & 33.24 & 28.17 & \textbf{46.46} & 28.21 & 10.91 \\
        & & RAMS & 30.34 & 38.03 & 38.01 & 23.76 & \textbf{53.07} & \underline{36.47} \\
        & & WikiEvents & 20.64 & 26.42 & 29.50 & 10.44 & 28.24 & 61.35 \\
        \midrule
        GPT-3.5 & \multirow{2}{*}{QA} & - & 27.56 &  25.60 & 18.54 & 26.69 & 22.54 & 10.36 \\ 
        GPT-4 & & - & 35.36 & 30.94 & 20.94 & \underline{36.29} & 24.10 & \phantom{0}7.09 \\ 
    \bottomrule
    \end{tabular}
    \caption{Argument $\text{F}_1$ for (1) TI and QA models fine-tuned and evaluated on all (source, target) dataset pairs (2) GPT-3.5 and GPT-4  evaluated zero-shot on all datasets. Best in-domain results are \textbf{bolded}; best zero-shot results are \underline{underlined}. Flan-T5 results are averages across three runs. GPT results are averages across three prompts.}
    \label{tab:results-flan-t5}
\end{table*}

Our experiments consider both (1) Flan-T5-based \citep{chung-etal-2022-scaling} QA and TI models fine-tuned and evaluated on all possible combinations of source and target dataset for our six datasets, and (2) GPT-3.5\footnote{\url{https://openai.com/blog/chatgpt}} and GPT-4 \citep{openai-2023-gpt4} evaluated zero-shot on each dataset. We briefly describe the datasets and models below.
\subsection{Data}
\label{experiments:data}
We use six datasets, each with its own ontology. All but ACE are document-level, i.e., arguments may appear in sentences other than the one containing their trigger. \autoref{app:datasets} has summary statistics.
\paragraph{ACE \cite{doddington-etal-2004-automatic}} is the most popular benchmark for sentence-level EAE, featuring a two-level event ontology of 33 types, along with 26 role types. The corpus contains news articles and broadcast transcripts covering business, financial transactions, military conflict, and life events. Sentences in ACE may contain multiple events.

\paragraph{ERE Light and Rich \citep{song-etal-2015-light}} were developed during the DARPA DEFT program \citep{darpa-2012-deft} as document-level successors to ACE and are based on its ontology, but differ in key ways. ERE-Light (ERE-L) and ACE each contain two event types the other does not, while ERE-L's 17 role types are a subset of the 26 in ACE. ERE-Rich (ERE-R) augments ERE-L with six further event types and nine further role types, and the resulting ontology overlaps with, but still differs from, ACE. ERE-L and ERE-R cover newswire and discussion forum documents, with only the latter annotated for both ontologies. Documents are annotated exhaustively for relevant events.

\paragraph{FAMuS \cite{vashishtha-etal-2023-famus}} is a dataset of short \emph{report} passages excerpted from Wikipedia articles, each paired with a corresponding (non-Wikipedia) \emph{source} article. A single event trigger is annotated in each report against a broad-coverage, 253-frame subset of the FrameNet ontology \citep[with 318 role types;][]{baker-etal-1998-berkeley} that contains only situation-denoting frames. Event arguments are annotated in both the report and the source, and may appear anywhere within either document. We use only the FAMuS reports in our experiments.

\paragraph{RAMS \cite{ebner-etal-2020-multi}} is a document-level dataset of news articles annotated against the three-level AIDA-I ontology with 139 event types and 65 role types. Like FAMuS, each document has one annotated event trigger, but arguments are limited to a five-sentence context window around it.

\paragraph{WikiEvents \cite{li-etal-2021-document}} covers web articles annotated against the three-level KAIROS ontology, with 49 event types and 57 role types. As in ERE, documents are exhaustively annotated.

\subsection{Models and Evaluation}
\label{experiments:models}
\paragraph{QA} We adopt a standard extractive QA architecture \citep{du-cardie-2020-event} that uses a Flan-T5-base encoder to jointly embed a question concatenated with the document context, containing a  highlighted event trigger. The embedding of the \texttt{BOS} token is then passed to a final linear layer to predict the start and end offsets of the answer span.\footnote{For roles with no arguments, we construct an example with target offsets $(0,0)$---i.e.\ the \texttt{BOS} token.} Since questions (roles) may have multiple answers (arguments), we construct a single example \emph{per argument} during training. At inference time, we construct a single example \emph{per role}, predicting up to $k$ argument spans for which model confidence exceeds a dev-tuned threshold (we set $k = 5$). We use the average of the cross-entropy losses w.r.t.\ the gold start and end offsets as the training objective.

\paragraph{TI} We draw on the approach of \citet{li-etal-2021-document} for our TI models. We use Flan-T5-base to (1) jointly encode a document containing a highlighted trigger together with its (unfilled) template, then (2) autoregressively decode the filled template. We use the NLL of the gold filled templates as the  objective and beam search for decoding ($\text{beam size} = 5$).

\paragraph{OpenAI Models} We evaluate GPT-3.5 and GPT-4 zero-shot, using the same questions as for our QA models. In the prompts, we provide instructions for the task and the same context passages with highlighted triggers as before. All questions are provided together in the same prompt, and we report averages across three prompt variants.\footnote{Further details on models and prompts in Apps.\ \ref{app:training} and \ref{app:openai}.}

\paragraph{Metrics} We report (typed) exact match argument $\text{F}_1$ for all settings. To allow direct comparisons among our models, we consider an argument to be correct iff it is string-equivalent to the reference.

\section{Results and Analysis}
\label{sec:discussion}
Results for all models can be found in \autoref{tab:results-flan-t5}. We present observations and further analysis below.

\paragraph{Effectiveness of Transfer with Small Models} We find that for each target ontology, \emph{some} Flan-T5 model (TI or QA) obtains zero-shot performance superior to GPT-3.5---often by wide margins. Remarkably, the same is also true w.r.t.\ GPT-4, with the lone exception of FAMuS. This is notable given the massive differences in the amount of pretraining data and in the parameter counts between Flan-T5-base (250\textbf{M}) and the GPT models (believed to be at least 175\textbf{B} for GPT-3.5 and at least 1.0\textbf{T} for GPT-4).\footnote{{\url{https://en.wikipedia.org/wiki/GPT-4}}} Even setting aside transfer between the most similar ontologies (ACE, ERE-\{R,L\}) and focusing on more distant ontology pairs, we \emph{still} observe stronger transfer results with some (indeed, for ERE-\{L,R\} and RAMS, even \emph{all}) Flan-T5 model(s) than with GPT-3.5. That said, transfer between distant pairs (e.g.\ WikiEvents $\rightarrow$ FAMuS) remains challenging in absolute terms. Given the cost and runtime of even the OpenAI \emph{inference} APIs, these results indicate that for many use cases, training a smaller model in-house on recasted existing resources can be a cheaper, more effective first-line approach to extraction in a new domain.

\begin{figure}
    \centering
    \includegraphics[width=\columnwidth]{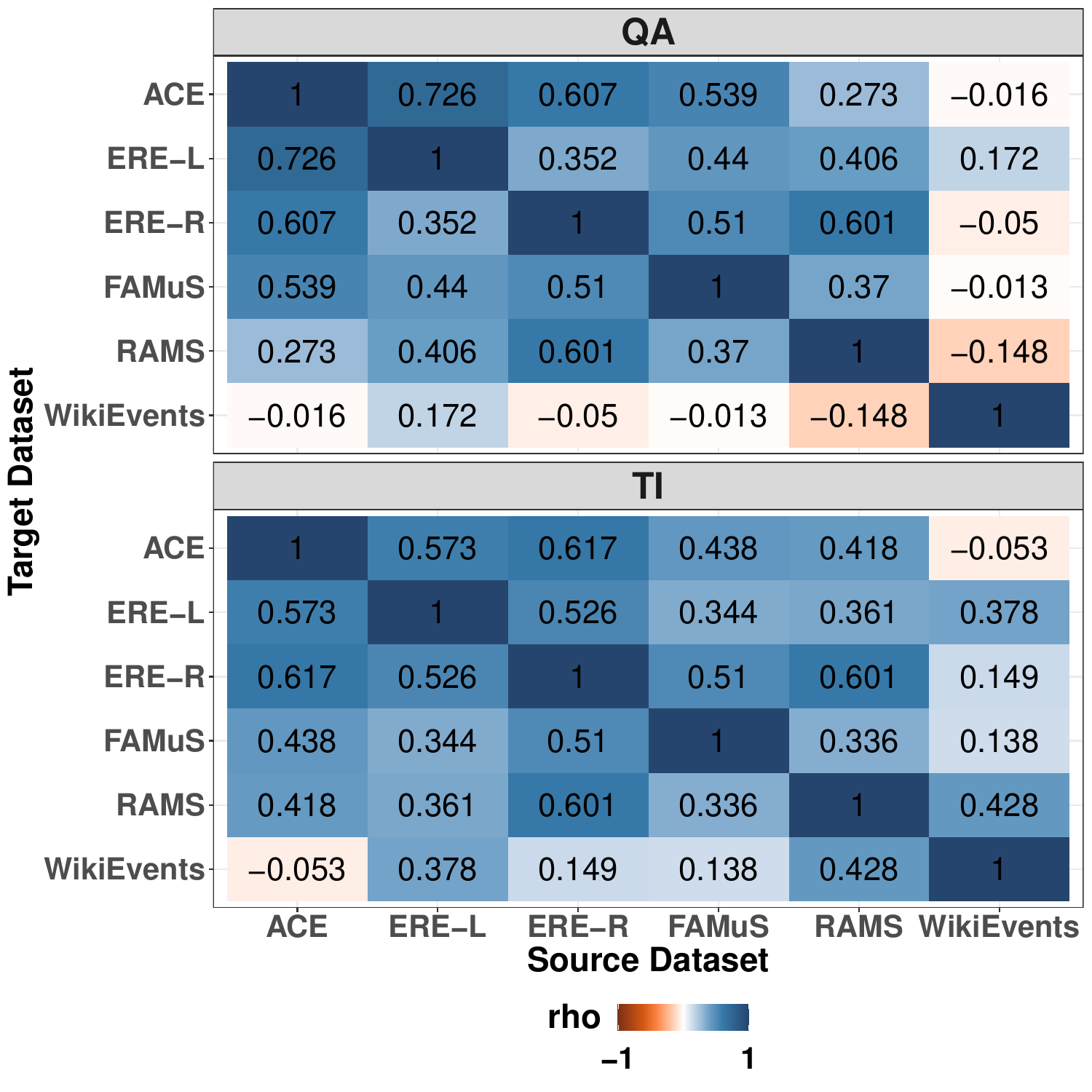}
    \caption{Pearson's $\rho$ between per-event type in-domain and zero-shot argument $\text{F}_1$.\vspace{-6mm}}
    \label{fig:event_type_correlations}
\end{figure}

\paragraph{TI vs.\ QA} We observe often sizable performance gaps between TI and QA models trained on the same source dataset, both in the in-domain evaluations (e.g.\ 65.95 on ACE for TI vs.\ 60.35 for QA) and and in the zero-shot evaluations (e.g.\ 25.37 on WikiEvents for RAMS-trained TI  vs.\ 36.47 for RAMS-trained QA). Yet, neither method is consistently dominant across domains: TI obtains best in-domain performance on 4/6 datasets, and best zero-shot performance on only 2/6. This suggests both TI and QA should be considered when attempting extraction on novel datasets and/or ontologies.

\paragraph{In-Domain vs.\ Out-of-Domain Performance} \autoref{fig:event_type_correlations} presents correlations ($\rho$) between in-domain and zero-shot argument $\text{F}_1$ at the event type level, with higher correlations indicating that the same event types tend to be hard/easy in both in-domain and zero-shot evaluations. We find higher correlations where the source and target ontologies are more similar (notably, among ACE and ERE-\{L,R\}), reflecting the smaller domain shifts.\footnote{In principle, this need not be the case---e.g.\ if the distributions of examples across types differed markedly.} WikiEvents stands out for its low (often negative) correlations, indicating little overlap in the types that are hard/easy in-domain vs.\ when transferring.

\begin{figure}
    \centering
    \includegraphics[width=\columnwidth]{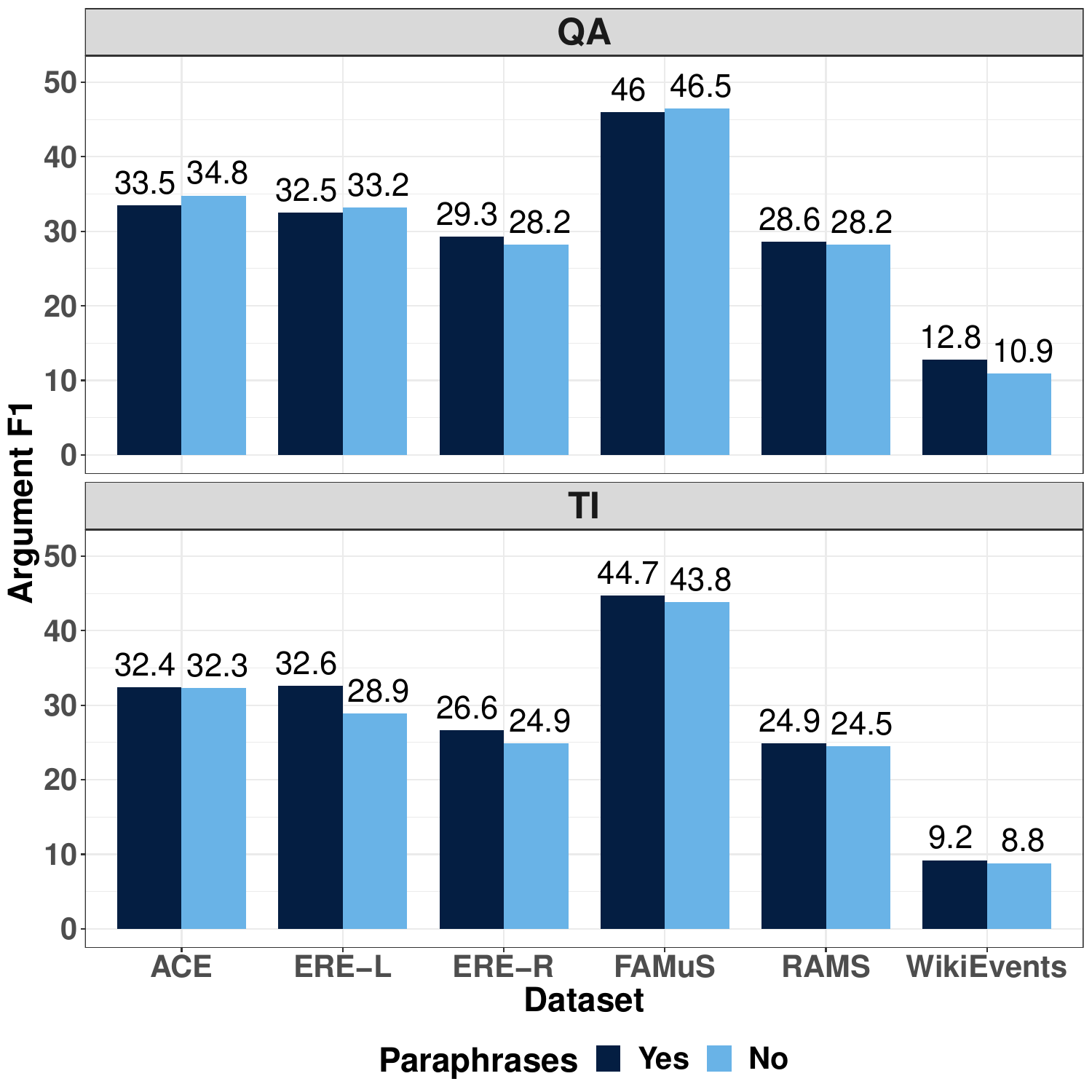}
    \caption{Argument $\text{F}_1$ of QA and TI models trained on FAMuS with  and without an additional five paraphrases per question/template.\vspace{-5mm}}
    \label{fig:paraphrase_results}
\end{figure}

\paragraph{Paraphrases} Lastly, we investigate the value of augmenting training data with \emph{paraphrases} of the questions and templates, focusing on transfer from FAMuS as a case study. We use GPT-4 to generate five paraphrases for each FAMuS question and template, and retrain the Flan-T5 QA and TI models on the augmented datasets.\footnote{GPT-4 prompts and hyperparameters are in \autoref{app:openai}.} \autoref{fig:paraphrase_results} reports average results across three runs. We find some gain from paraphrases across all ontologies in the TI setting (from 0.1 $\text{F}_1$ on ACE, up to 3.7 $\text{F}_1$ on ERE-L), with more mixed results in the QA setting (modest gains for ERE-R, RAMS, and WikiEvents only). Paraphrases thus may (not) be worthwhile depending on the size of the gains and one's training budget.

\section{Conclusion}
\label{sec:conclusion}
This work has investigated two prominent methods for EAE  transfer learning---question answering (QA) and template infilling (TI)---across six datasets. While neither method consistently outperforms the other, we have shown that small models trained with these methods on an appropriate source ontology can yield zero-shot extraction performance superior to that of vastly larger models (GPT-3.5, GPT-4). At least for EAE, our results suggest that, far from being obsolete, small models trained on recasted existing resources remain an effective first choice for extraction in new domains.

\section*{Limitations}
\label{sec:limitations}
First, this work has focused on EAE, which uses gold event triggers. It is possible findings could differ in the full event extraction setting. Such an investigation would be an interesting direction for future work, though we note that it would be difficult to carry out with the same set of datasets considered here, as some (RAMS and FAMuS) are not exhaustively annotated for relevant events, creating challenges around censored data.

Second, we use the term \emph{zero-shot} to refer to settings in which a model is evaluated on a different dataset/ontology from that on which it was trained. However, there is some (variable) overlap in the event and role types in the ontologies we consider. We do not think this impugns the import of our findings, since most practical EAE transfer scenarios are ones in which at least some key features of the source domain are at least partially preserved in the target (e.g.\ agent-like and patient-like roles may be expected to exist in both). However, it is possible we would find poorer transfer results for yet more divergent domain pairs than those considered here.

Finally, transfer performance for most ontology pairs remains low in absolute terms, which may present an obstacle to deploying these methods in some real-world scenarios.

\section*{Ethics}
\label{sec:ethics}
As this work primarily evaluates existing models on existing resources, we do not believe it introduces any novel ethical concerns. However, many of the corpora studied in this work discuss historical situations involving military and political violence, and caution is therefore warranted in using them to train models for real world applications.

\bibliography{anthology,custom}
\bibliographystyle{acl_natbib}

\clearpage
\appendix

\section{Dataset Details}
\label{app:datasets}
\begin{table}
\centering
\adjustbox{max width=\columnwidth}{%
\begin{tabular}{l|lllll}
\toprule
    Dataset & Event Types & Role Types & Events & Arguments & Doc Level? \\
\midrule
    ACE & 33 & 26 & 5,223 & 9,629 & No \\
    ERE-L & 33 & 17 & 4,066 & 5,474 & Yes \\
    ERE-R & 38 & 26 & 5,763 & 10,621 & Yes \\
    FAMuS & 253 & 318 & 1,265 & 3,953 & Yes \\
    RAMS & 139 & 65 & 9,124 & 21,237 & Yes \\
    WikiEvents & 49 & 57 & 3,951 & 5,536 & Yes \\
\bottomrule
\end{tabular}}
\caption{Statistics for all datasets considered in this work. Counts reflect totals over all datasets. Identically named role types for different event types are treated as the same role type for counting purposes. FAMuS statistics are for the report documents only.}
\label{tab:dataset_statistics}
\end{table}

\subsection{Summary Statistics}
All the data used in this work is English only. \autoref{tab:dataset_statistics} contains summary statistics for all of the datasets/ontologies considered in this work. Counts reflect totals over all splits.

\subsection{Licenses}
ACE, ERE-L, and ERE-R are available through the Linguistic Data Consortium (LDC; ACE Catalog No.: LDC2006T06; ERE Catalog No.: LDC2023T0). These are licensed under the \href{https://catalog.ldc.upenn.edu/license/ldc-non-members-agreement.pdf}{LDC User Agreement for Non-Members}, which allows for use of LDC artifacts ``only for non-commercial linguistic education, research and technology development.'' Our use of these datasets is for research purposes only and thus adheres to the license.

We use \href{https://nlp.jhu.edu/rams/}{RAMS v1.0c} and \href{https://github.com/FACTSlab/FAMuS/commit/4c8cc1f1f09d859940992016061f6cb9686c573d}{FAMuS} (unversioned; commit=4c8cc1f), which are both released under a \href{https://creativecommons.org/licenses/by-sa/4.0/}{CC-BY-SA-4.0} License. This license allows one to ``remix, transform, and build upon the material for any purpose, even commercially,'' provided the resulting work contains proper attribution to the original and is released under the same license. We do not alter the RAMS or FAMuS datasets in any way, though our experiments use (and acknowledge) them, and our code will be released under a CC-BY-SA-4.0 license, which is consistent with the terms of use.

We use the version of WikiEvents released at \url{https://github.com/raspberryice/gen-arg/} (unversioned; commit=253e088). The only license provided is an MIT License, covering the repository as a whole, which grants permission to ``to use, copy, modify, merge, publish, distribute, sublicense, and/or sell copies of the Software.'' We believe our use of WikiEvents is thus consistent with this license.

\section{Implementation Details}
\label{app:training}
All Flan-T5 models (both TI and QA) are initialized from the \texttt{google/flan-t5-base} pretrained model and are trained using the HuggingFace Transformers package \citep[v4.38.2;][]{wolf-etal-2019-huggingface}, which relies on PyTorch \citep[v2.0.1;][]{paszke-etal-2019-pytorch}. We use the HuggingFace Tokenizers package (v0.15.2; \emph{ibid.}) and SpaCy \citep[v3.7.4;][]{spacy2} for tokenization.\footnote{Preprocessing details are available in our GitHub repo: \tiny{\url{https://github.com/wgantt/eae-transfer}}.}Evaluation is performed using the Metametric package \citep[v0.1.0-alpha.4;][]{chen-etal-2023-unified}. We train for a maximum of 50 epochs using the AdamW optimizer \citep{loshchilov-hutter-2018-decoupled} with default hyperparameters and full precision on a combination of NVIDIA RTX 6000 and A100 GPUs (CUDA v11.7).  We use argument $\text{F}_1$ on the dev split as the stopping criterion, with a patience of 10 epochs. The only hyperparameter tuning we perform is manual tuning of the batch size. For the QA models, batch size was set to 32 (except for the FAMuS models, where it was set to 4). For the TI models, batch size was set to 8. Results in \autoref{tab:results-flan-t5} reflect averages across three runs using random seeds 1337, 1338, and 1339 set globally. GitHub Copilot was used to assist in coding.

To choose the confidence threshold used for argument selection in the QA models, we sweep  thresholds in increments of 5\% from the $5^\text{th}$ to the $95^\text{th}$ percentiles of confidence scores over all $k \cdot \lvert\mathcal{D}_\text{dev}\rvert$ candidate arguments for the dev split, where $k$ is a hyperparameter that determines the maximum number of arguments that may be predicted for a given role and where $\lvert\mathcal{D}_\text{dev}\rvert$ is the total number of examples (=role instances) in the dev split. At the end of each epoch, we select the threshold that yields the highest dev argument $\text{F}_1$. For test set evaluation, we use the dev threshold $t_\text{dev}$ of the checkpoint with highest dev argument $\text{F}_1$. The same threshold is used for all datasets at test time time. Finally, we require that model confidence for all predicted arguments exceeds the confidence in the ``no answer'' response (i.e.\ confidence in the \texttt{BOS} token). Thus if the model confidence for ``no answer'' is $c_\text{null}$, the final threshold is given by $\text{max}(t_\text{dev}, c_\text{null})$.

\section{OpenAI}
\label{app:openai}
\subsection{Main Experiments}
We use the \texttt{gpt-3.5-turbo-0125} version of GPT-3.5 and the \texttt{gpt-4-0125-preview} version of GPT-4. For both, we set $\texttt{top\_p}=1.0$, $\texttt{temperature}=0.7$, $\texttt{max\_new\_tokens} = 512$, and we use no frequency or presence penalties. Our three prompts each use a different system prompt, listed below. The corresponding user prompts are described further down. Prompts and hyperparameters were chosen based on their promising results in manual prompt engineering efforts on the OpenAI playground that used several training set examples.\footnote{\url{https://platform.openai.com/playground}}

\paragraph{System Prompts}
\begin{enumerate}
\item You are the world champion of extractive question answering.
\item You are an expert at question answering from text.
\item You are the best in the world at reading comprehension.
\end{enumerate}

\noindent Our three user prompts share the same basic format (see below), but vary in the wording of their instructions. Complete (system+user) prompts are constructed by pairing system prompt $i \in \{1,2,3\}$ with a user prompt containing instruction set $i$. As indicated in the \textbf{instructions}, each $\langle\texttt{Document}\rangle$ contains a single highlighted event trigger.

\paragraph{User Prompt Format\\}

\noindent Instructions: $\langle\texttt{Instructions} \rangle$
You must give your answers as JSON in the following format:
\begin{lstlisting}
{
  "q1": [ ... ],
  ...,
  "qN": [ ... ]
}
\end{lstlisting}
Input Passage: $\langle\texttt{Document}\rangle$

\noindent Answers:

\paragraph{Instructions}
\begin{enumerate}
\item I will give you an input passage containing an event trigger demarcated with “<trigger></trigger>” HTML tags. I will also give you a set of questions about the event denoted by that trigger. Your task is to answer each question with a list of contiguous spans extracted from the input passage. Answers may contain zero, one, or multiple spans. The list should be empty if no answer can be found.
\item I will show you a document that contains an event trigger that is highlighted with “<trigger></trigger>” HTML tags. After the document, I will list out a set of questions about the event referred to by the highlighted trigger. Please answer each question with a list of zero or more contiguous spans extracted from the input passage. Spans MUST appear in the document. Some questions may not have answers, in which case the answer should be an empty list.
\item I will give you a passage of text featuring a phrase that refers to some event and that is highlighted with '<trigger></trigger>' HTML tags. I will additionally provide you with a list of questions about the event referred to by the highlighted phrase. You must answer each question with a list of zero, one, or multiple contiguous spans that appear in the input passage. Some questions do not have any answer in the input passage. For these cases, your answer should be an empty list.
\end{enumerate}

\subsection{Paraphrase Generation}
To generate paraphrases for the experiments in \S\ref{sec:discussion}, we use \texttt{gpt-4-0125-preview} with $\texttt{top\_p} = 1.0$, $\texttt{temperature} = 0.7$, and $\texttt{max\_new\_tokens} = 512$ with no system prompt and with the user prompts shown below (different for questions and for templates). 

\paragraph{Template Prompt\\}
Instructions: Please generate five paraphrases of the following template, but you ABSOLUTELY CANNOT change any words that are in between brackets ([]). Your paraphrases MUST be formatted as a JSON list of strings.\\~\\
Template: $\langle \texttt{template} \rangle$\\~\\
Paraphrases:

\paragraph{Question Prompt\\}
Instructions: Please generate five paraphrases of the following question. Your answer MUST be formatted as a JSON list of strings.\\~\\
Question: $\langle \texttt{question} \rangle$\\~\\
Paraphrases:


\end{document}